\documentclass{article}
\usepackage{spconf,amsmath,graphicx}
\usepackage{epsfig}
\usepackage{graphicx}
\usepackage{amsmath}
\usepackage{amssymb}
\usepackage{algorithm}
\usepackage{algpseudocode}
\usepackage{url}
\usepackage[tight,footnotesize]{subfigure}
\usepackage{multirow}
\usepackage{array}
\usepackage{color}
\usepackage{bm}
\usepackage[normalem]{ulem}

\title{Novel Evaluation Metrics for Seam Carving based Image Retargeting}
\twoauthors
  {Tam V. Nguyen}
	{Department of Computer Science\\
	University of Dayton\\
Email: tamnguyen@udayton.edu}
  {Guangyu Gao}
	{School of Software\\Beijing Institute of Technology\\
	Email: guangyugao@bit.edu.cn}
\begin{document}
%
\maketitle
\begin{abstract}
Image retargeting effectively resizes images by preserving the recognizability of important image regions. Most of retargeting methods rely on good importance maps as a cue to retain or remove certain regions in the input image. In addition, the traditional evaluation exhaustively depends on user ratings. There is a legitimate need for a methodological approach for evaluating retargeted results. Therefore, in this paper, we conduct a study and analysis on the prominent method in image retargeting, Seam Carving. First, we introduce two novel evaluation metrics which can be considered as the proxy of user ratings. Second, we exploit salient object dataset as a benchmark for this task. We then investigate different types of importance maps for this particular problem. The experiments show that humans in general agree with the evaluation metrics on the retargeted results and some importance map methods are consistently more favorable than others.  
\end{abstract}
\linespread{0.955}

\section{Introduction}

Image retargeting, sometimes referred as image cropping, thumbnailing, or resizing, is beneficial for some practical scenarios, \textit{i.e.}, facilitating large image viewing in small size displays, particularly on mobile devices. This is a very challenging task since it requires preserving the relevant information while maintaining an aesthetically pleasing image for viewers. The premise of this task is to remove indistinct regions and retain the context with the most salient regions. In the pioneering work, Setlur \textit{et al.}~\cite{Setlur} propose using an importance map of the source image obtained from saliency and face detection. In the importance map, the pixels with higher values are most likely preserved and vice versa. If the specified size contains all the important regions, the source image is simply cropped. Otherwise,  the important regions are removed from the image, and fill the resulting ``holes'' using the background creation technique. Later, Avidan~\textit{et al.}~\cite{SeamCarving} propose the Seam Carving method based on the importance map computed from gradient magnitude. Seam Carving functions by constructing a number of \textit{seams} (paths of least importance) in an image and automatically removes seams to reduce image size. Zhang \textit{et al.}~\cite{ZhangMingMing} present an image resizing method that attempts to ensure that important local regions undergo a geometric similarity transformation, and at the same time, image edge structure is preserved. Suh \textit{et al.}~\cite{Suh} propose a general thumbnail cropping method based on a saliency model that finds the informative portion of images and cuts out the non-core part of images. Marchesotti \textit{et al.}~\cite{Marchesotti} propose a framework for image thumbnailing based on visual similarity. Their underlying assumption is that images sharing their global visual appearance are likely to share similar saliency values. While other works are dedicated to still images, Chamaret and Le Meur~\cite{Chamaret} propose a video  retargeting algorithm. Meanwhile, Rubinstein~\textit{et al.}~\cite{Rubinstein} extend Seam Carving~\cite{SeamCarving} into  video retargeting.

\begin{figure}[!t]
\centering
\includegraphics[width = \linewidth]{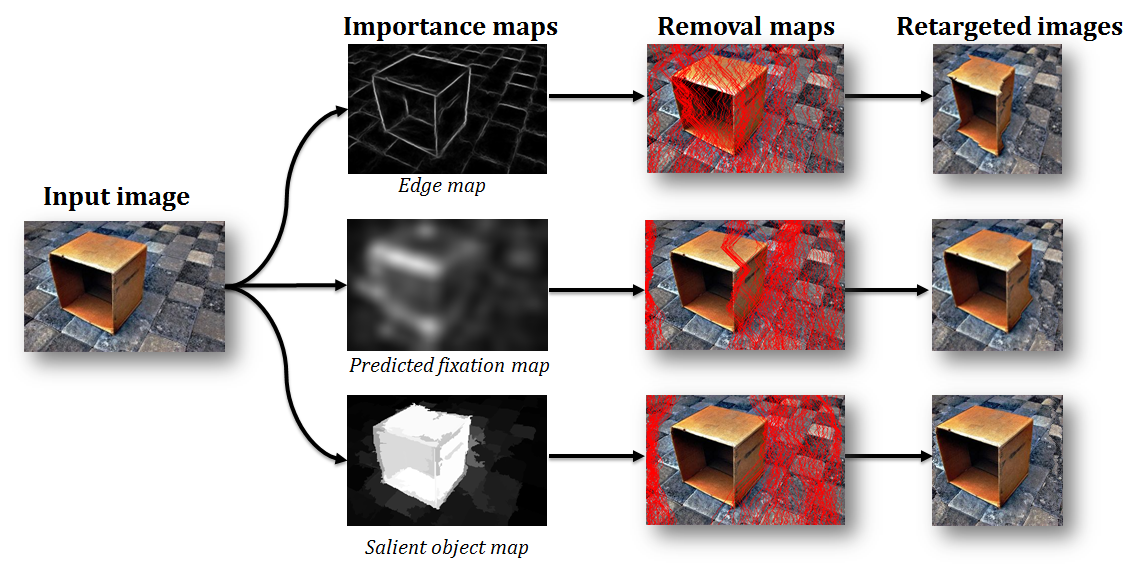}
\caption{The flowchart of Seam Carving on a given image with the importance map from different methods, namely, edge detector, human fixation predictor, and salient object detector. The removal map is later generated by highlighting the least important seams. The red lines are represented the removal seams. The accordingly retargeted images are finally constructed by removing the red lines to reach the desired size.} 
\label{fig:squareme}
\end{figure}

\begin{figure*}[!t]
\centering
\includegraphics[width = \linewidth]{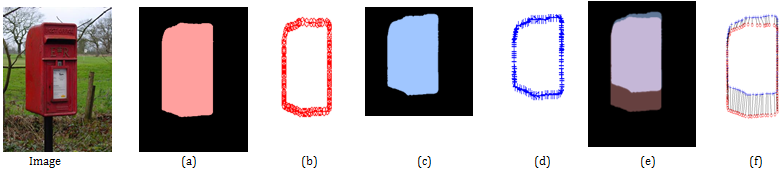}
\caption{The two novel metrics, namely, Mean Area Ratio, and Mean Sum of Squared Distances. From left to right: Original image, (a) the ground truth saliency map, (b) the shape points of the ground truth map, (c) the retargeted ground truth map from COV~\cite{COV}, (d) the shape points of the retargeted ground truth map, (e) the mean area ratio map, (f) the mapping between two correspondence sets. 
}
\label{fig:metrics}
\end{figure*}

To date, the existing evaluation scheme mostly depends on user ratings. However, it is not always feasible to recruit a large pool of participants for the evaluation. Also, there is mostly impossible to get the same participant pool of a previous work to make a fair comparison. Thus there is a legitimate need of an automatic way to evaluate these retargeting methods. In this paper, we revisit and further analyze the most popular method, Seam Carving, for image retargeting. Our contribution is two-fold. First, we propose two novel metrics to systematically evaluate the retargeting algorithms, namely, Mean Area Ratio (MAR) and Mean Sum of Squared Distances (MSSD). Our novel metrics focus on how much shape of the salient object(s) is distorted after the retargeting process. Second, we evaluate various types of importance map, namely, fixation prediction map, salient object map, and edge map, with the newly proposed metrics.

\section{Seam Carving Revisit and Proposed Evaluation Metrics}

\subsection{Seam Carving Revisit}
\label{retargeting}
Seam Carving, the most popular method in image retargeting, aims to automatically retarget the images into a certain size to facilitate the viewing purpose as aforementioned. Let $\bm{I}$ be an $m \times n$ image. As illustrated in Figure \ref{fig:squareme}, the first step is the computation of an importance map $\bm{S}$, which quantifies the importance of every pixel in the image. Every pixel in the importance map is assigned a value within $[0, 1]$, where higher values mean higher importance. Assume $\bm{I}$ is a landscape image where $n > m$, we aim to reduce its width. The vertical seam $s$, an 8-connected path in the image from the top to the bottom containing one pixel per row, is defined as below:

\begin{equation}
	s = \{s_i\}_{i=1} ^ {m} = \{(i, y(i))\}_{i=1} ^ {m} , s.t. \forall i,|y(i) - y(i-1)| \leq 1,
\end{equation}

where $y(i)$ is the corresponding column of row $i$ within the seam. Our goal is to find the optimal seam that minimizes:
\begin{equation}
\label{eq:seam}
s^{*} = \min_s{\sum_{i=1}^m \bm{S}(s_i)},
\end{equation}
where $\bm{S}(s_i)$ is the importance value of one seam pixel. Eqn.~(\ref{eq:seam}) can be solved by dynamic programming. This optimal seam is later removed out of the input image. This process repeats until the image reaches its desired dimension.

It is worth noting that the recent years witness the rapid popularity of smartphones and tablets that equips people with imaging capabilities. In fact, people are taking photos in different ways. Traditional filmmakers take more photos about the landscape than human figures. However, on a mobile phone, people prefer to take pictures in the portrait mode. Due to this difference in people's preferences, applications like Instagram have been developed which meets the demands of both groups of people by asking them to crop the image to the square size. In the social media, most of profile images are in the square form, \textit{i.e.}, Facebook and Twitter. One reasonable explanation is that squared photos display well in a feed format. In this work, we utilize the Seam Carving method into an application, so called \textbf{Make-It-Square}, which automatically retargets images into the square size.  In particular, the Seam Carving process loops for $n - m$ times until the landscape image reaches its expected square size. For the portrait image, we transpose the image and use the same function to find the optimal vertical seam.

\begin{figure*}[!t] 
\centering \includegraphics[width = \linewidth]{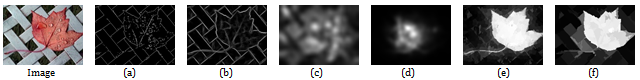} 
\caption{From left to right: Original image, and importance maps from 6 different methods: (a) Sobel edge map~\cite{Sobel}, (b) Structured edge map~\cite{Dollar}, (c) boolean map based saliency (BMS~\cite{BMS}), (d) saliency based on region covariance (COV~\cite{COV}), (e) high-dimensional color transform (HDCT~\cite{HDCT}), (f) discriminative regional feature integration (DRFI~\cite{DRFI}).}  
\label{fig:saliency} 
\end{figure*}

\subsection{Proposed Evaluation Metrics}

In order to mitigate the dependency of user ratings, we propose two additional metrics to systematically evaluate the retargeting algorithms, namely, Mean Area Ratio and Mean Sum of Squared Distances. Our motivation is that the users prefer the shape of the salient object(s) is preserved after the image retargeting process as discussed in~\cite{SeamCarving}. As shown in Fig.~\ref{fig:squareme}, the distorted boxes in the first two rows (retargeted images) are not entertained by the viewers.

Our first metric, the Mean Area Ratio, measures how much the salient object(s) can be preserved after the image retargeting. We simultaneously remove seams on both the original image and its ground truth saliency map. Obviously, the retargeted groundtruth map has the exactly same size with the retargeted image. For each input image, the area ratio is computed as the ratio between the salient regions in the retargeted ground truth map and the ground truth salient areas, as shown in Fig.~\ref{fig:metrics}e. The area ratio is $1$ when the whole salient regions are retained. The Mean Area Ratio, MAR, for a set of input images is computed over the area ratios of all images. 

Our second metric, the Mean Sum of Squared Distances, evaluates the shape similarity of the salient regions before and after the image retargeting. We adopt Shape Contexts~\cite{ShapeContext} to measure the shape similarity. For each image, Shape Contexts compute the shape correspondences of two given silhouettes (the ground truth map and the retargeted ground truth maps as shown in Fig.~\ref{fig:metrics}b, d). Next, the distances between two correspondence sets are summed as illustrated in Fig.~\ref{fig:metrics}f. The sum of squared distances is $0$  when two shapes are identical. Eventually the Mean Sum of Squared Distances, MSSD, is computed across over all images. 

Actually, the two proposed evaluation metrics are complementary to each other. MAR measures how much salient object(s) are maintained, whereas MSSD measures the amount of distortion after the image retargeting process.

\subsection{Selection of Importance Map}

In literature, the edge map is first introduced as the importance map for image retargeting problem~\cite{SeamCarving}. Additionally, the importance level can be measured by visual saliency values. There exist two popular outputs of visual saliency prediction, namely, the predicted human fixation map for fixation prediction, and the salient object map for salient object/region detection. In literature, there also exist many efforts to predict visual saliency with different cues, \textit{i.e.}, depth matters~\cite{Depth}, audio source~\cite{Audio}, touch behavior~\cite{Touch},  object proposals~\cite{AH1,AH}, and semantic priors~\cite{SP}. In this paper, we consider three types of importance maps as follows. 

\textbf{Edge map} is retrieved from the edge detection process, a fundamental task in computer vision since the early 1970's~\cite{Duda,Robinson}. Early works~\cite{Sobel,Canny} focused on the detection of intensity or color gradients. For example, the popular Sobel detector~\cite{Sobel} computes an approximation of the gradient of the image intensity function. Recently, Dollar \textit{et al.}~\cite{Dollar} proposed structured edge detection (SE) by formulating the problem of edge detection as predicting local segmentation masks given input image patches.  In this work, we consider different edge detectors~\cite{Sobel,Dollar}.

\textbf{Fixation prediction map} is obtained from trained models which are constructed originally to understand
human viewing patterns. Actually, these models aim to predict points that people look at (freeviewing of natural scenes usually for 3-5 seconds). The typical ground-truth fixation map includes several fixation points smoothened by a Gaussian kernel. We consider using two state-of-the-art models, namely, Boolean Map based Saliency (BMS~\cite{BMS}) and saliency based on region covariance (COV~\cite{COV}) for the later evaluation. 

\textbf{Salient object map} is computed from models which
aim to detect and segment the most salient object(s) as a
whole. Note that a typical pixel-accurate ground-truth  map usually contains several regions marked by humans. As recommended in the extensive survey~\cite{BorjiReview}, we consider two state-of-the-art models, namely, saliency based on Discriminative Regional Feature Integration (DRFI~\cite{DRFI}) and High-Dimensional Color Transform (HDCT~\cite{HDCT}).

Fig.~\ref{fig:saliency} shows the importance maps generated from different computational methods. Note that edge maps and fixation prediction maps are of low resolution and highlight edges whereas the salient object maps focus on the entire objects.

\begin{figure*}[!t]
\centering
\includegraphics[width = \linewidth]{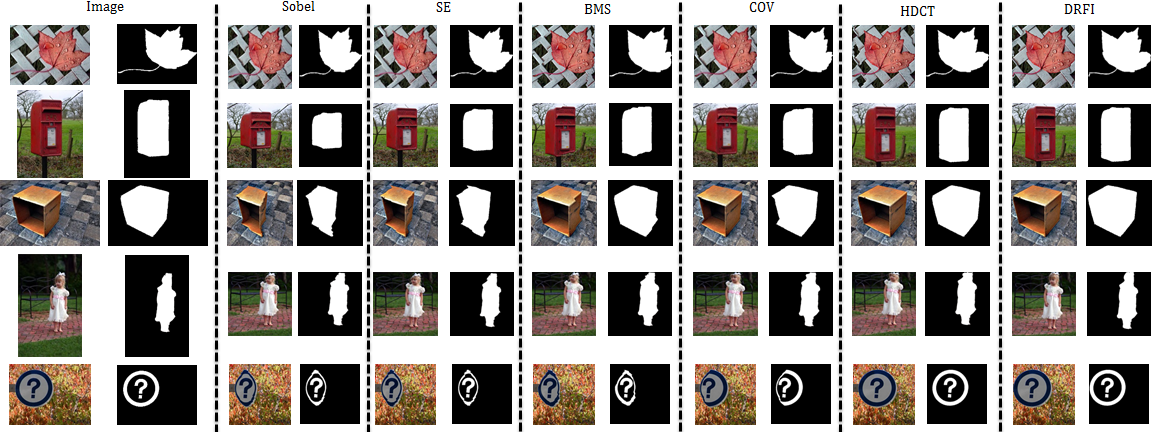}
\caption{Visual comparison of retargeted images from different importance maps on MSRA-1000 dataset~\cite{FT}. From left to right: Original image, the ground truth saliency map, the pairs of retargeted image and the retargeted groundtruth saliency map with the importance maps from Sobel, Structured Edge (SE), BMS, VOC, HDCT, DRFI, respectively. (Please view in high 400\% resolution for best visual effect).}
\label{fig:retarget}
\end{figure*}

\section{Evaluation}

It is obvious that the benchmark of image retargeting task requires a set of input images with their corresponding saliency map. This requirement elegantly fits the settings of salient object datasets. Therefore, we exploit the popular MSRA-1000 dataset \cite{FT}, which contains $1,000$ images with the annotated pixel-wise ground truth of salient regions, for the evaluation.

We first show the visual comparison of retargeted images from different importance maps. As observed from Fig.~\ref{fig:retarget}, the retargeted results from salient object detection methods well preserve the main salient objects without distortion. Though fixation prediction is in general biologically plausible and suggests important regions as the way as humans look at, their retargeted images lose details. Meanwhile, the retargeted images from edge-based importance map lose both details and layout structure. 

Next, we conduct a user study to evaluate the performance of retargeted images from different input saliency maps on previously mentioned MSRA-1000 dataset \cite{FT}. We run Make-It-Square on the dataset to obtain $1,000$ retargeted squared images. 40 participants (14 are female) who are university staff/students are involved in this experiment, and a set of images is provided to each participant. Note that every image set contains $50$ random images and six other retargeted results where each method is randomly labeled from 1 to 6 to hide identities. The participant is requested to rate all methods  with the scores (1-6), where 1 means bad viewing experience and 6 means excellent viewing experience. As shown in Table~\ref{table:eval}, users prefer the salient object map methods, HDCT~\cite{HDCT} and DRFI~\cite{DRFI}, whereas the retargeted results from edge map, Sobel~\cite{Sobel}, Structured Edge~\cite{Dollar}, receive the least rating.

\begin{table}[]
\centering
\caption{The performance of different importance maps on image retargeting.}
\label{table:eval}
\small
\begin{tabular}{|l|c|c|c|}
\hline
\textbf{Importance Map} & \textbf{User Ratings }& ~~~~\textbf{MAR}~~~~    & ~\textbf{MSSD}~   \\ \hline
Sobel~\cite{Sobel} &1.3         & 0.8976 & 0.0406 \\ \hline
Structured Edge~\cite{Dollar}  &1.9       & 0.9132 & 0.0402 \\ \hline
COV~\cite{COV}    &3.5        & 0.9581 & 0.0395 \\ \hline
BMS~\cite{BMS}    &3.2        & 0.9638 & 0.0395 \\ \hline
HDCT~\cite{HDCT}    &5.4       & 0.9840 & 0.0387 \\ \hline
DRFI~\cite{DRFI}     &5.7      & 0.9877 & 0.0389 \\ \hline
\end{tabular}
\end{table}

We then compute two evaluation metrics, MAR and MSSD, and the results are generally similar with user ratings. Also shown in Table~\ref{table:eval}, the retargeted images obtained from the salient object map source are consistently more favorable than others, namely, achieving the highest MAR and the lowest MSSD. On the contrary, the retargeted results of edge maps receive the lowest MAR and the highest MSSD. 

In addition, we further compute the Pearson coefficient correlations (CC) (defined in~\cite{BorjiReview}) between user ratings and the two novel metrics. Note that the correlation of one metric score and itself is $1$. As shown in Table~\ref{table:cc}, the CCs between user ratings and MAR and negative MSSD are $0.955$ and $0.977$, respectively. This demonstrates those two metrics are highly correlated with users's responses. Hence, the proposed metrics can be used as the proxy of user ratings.

\begin{table}[]
\centering
\caption{The Pearson coefficient correlation ~\cite{BorjiReview} among three metrics, user ratings, MAR and MSSD.}
\label{table:cc}
\small
\begin{tabular}{|l|c|c|c|}
\hline
 & {User Ratings}& ~~~~{MAR}~~~~~    & ~ {- MSSD}~   \\ \hline
{User Ratings}  &1       & 0.955 & 0.977 \\ \hline
{MAR} &0.955         & 1 & 0.981 \\ \hline
{- MSSD}    &0.977        & 0.981 & 1 \\ \hline
\end{tabular}
\end{table}

\section{Conclusion and Future Work}
In this paper, we introduce two novel metrics to automatically evaluate Seam Carving for the image retargeting task. We utilized salient object dataset as a benchmark and showed that the newly proposed metrics are highly correlated with the user ratings across six different importance maps. We also found that the retargeted results, with the salient object map used as the importance map, are consistently more favorable than others. We believe that the new benchmark type and our evaluation measures will lead to improved retargeting algorithms, as well as better understanding of image retargeting problem.

For future work, we aim to investigate other image retargeting operators apart from Seam Carving. We also would like to extend our work by considering additional cues, \textit{e.g.}, the depth in RGBD images or motion information in videos. 

\bibliographystyle{IEEEbib}
\bibliography{sigproc2} 
\end{document}